\documentclass[journal,12pt,onecolumn,draftclsnofoot,]{IEEEtran}
\IEEEoverridecommandlockouts
\usepackage{cite}
\usepackage{amsmath,amssymb,amsfonts}
\usepackage{algorithmic}
\usepackage{graphicx}
\usepackage{blindtext}
\usepackage{hyperref}
\usepackage{textcomp}
\usepackage[dvipsnames]{xcolor}
\usepackage{cite}
\usepackage[colorinlistoftodos]{todonotes}
\usepackage{amsmath,amssymb,amsfonts}
\usepackage{algorithm}
\usepackage{multirow}
\usepackage{amsmath}
\usepackage[bb=dsserif]{mathalpha}
\usepackage{bm}
\usepackage{bbm}
\usepackage{siunitx}
\usepackage{algorithmic}
\usepackage{comment}
\usepackage{caption}
\usepackage{subcaption}
\usepackage{booktabs,
            makecell, 
            tabularx}
\usepackage{textcomp}
\usepackage{mathtools}
\def\BibTeX{{\rm B\kern-.05em{\sc i\kern-.025em b}\kern-.08em
    T\kern-.1667em\lower.7ex\hbox{E}\kern-.125emX}}

\ifCLASSINFOpdf

\else

\fi

\begin{document}

\title{DySurv: Dynamic Deep Learning Model for Survival Analysis with Conditional Variational Inference}

\author{Munib Mesinovic, MSc\textsuperscript{1}, Peter Watkinson, PhD\textsuperscript{2}, Tingting Zhu, PhD\textsuperscript{1}
\thanks{\textsuperscript{1} Department of Engineering Science, University of Oxford, Oxford, UK, e-mail: munib.mesinovic@jesus.ox.ax.uk}
\thanks{\textsuperscript{2} Critical Care Research Group, Nuffield Department of Clinical Neurosciences, University of Oxford, Oxford, UK}
}
\maketitle

\begin{abstract}

\noindent 

\textbf{Objective:} Machine learning applications for longitudinal electronic health records often forecast the risk of events at fixed time points whereas survival analysis achieves dynamic risk prediction by estimating time-to-event distributions. Here, we propose a novel conditional variational autoencoder-based method, DySurv, which uses a combination of static and longitudinal measurements from electronic health records to estimate the individual risk of death dynamically.\\

\textbf{Materials and Methods:} DySurv directly estimates the cumulative risk incidence function without making any parametric assumptions on the underlying stochastic process of the time-to-event. We evaluate DySurv on 6 time-to-event benchmark datasets in healthcare, as well as two real-world ICU EHR datasets extracted from eICU and MIMIC-IV. \\

\textbf{Results:} DySurv outperforms other existing statistical and deep learning approaches to time-to-event analysis across concordance and other metrics. It achieves time-dependent concordance of over 60\% in the eICU case. It is also over 12\% more accurate and 22\% more sensitive than in-use ICU scores like APACHE and SOFA. The predictive capacity of DySurv is consistent and the survival estimates remain disentangled across different datasets. \\

\textbf{Discussion:}  Our interdisciplinary framework successfully incorporates deep learning, survival analysis, and intensive care to create a novel method for time-to-event prediction from longitudinal health records. We test our method on several held-out test sets from a variety of healthcare datasets and compare it to existing in-use clinical risk scoring benchmarks. \\

\textbf{Conclusion:} While our method leverages non-parametric extensions to deep learning-guided estimations of the survival distribution, further deep learning paradigms could be explored.
\end{abstract}


\IEEEpeerreviewmaketitle

\section{BACKGROUND}
\noindent Survival analysis refers to statistical approaches to estimating distributions of event times or times it takes for an event to happen as well as rates of survival over time while accounting for censoring. The events in question can be machine failures in industry or the occurrence of specific diseases and death \cite{lee2006threshold}. In clinical practice, survival analysis can play a key role and provide valuable information to predict patient outcomes and guide treatment decisions \cite{yoon2017personalized}. While most traditional applications of survival analysis occur in epidemiology at the population level, with the rise of deep learning techniques, personalised estimation of survival times for individual patients has become possible \cite{luck2017deep}. However, the limitations of standard statistical models such as the Cox proportional hazards model include lack of complexity, constraining assumptions about the behaviour or proportions of covariate effects over time, and the reliance on only using static covariates \cite{zhong2021deep}. While there are extensions of Cox to time-varying covariates, these still suffer from the proportionality assumption (often violated in practice) and are outperformed by simple deep learning models \cite{kvamme2019time}. The proportional hazards assumption inherent in Cox models states that the hazard rates of different patients over time remain in fixed proportion to each other, in other words, despite the potential change in risk trajectories for a patient in the future, these changes remain fixed relative to other patients. Thus, implying that the hazard in measuring the effect of any predictor is constant over time. This severely restraints the predicted risk trajectories and they often do not capture real changes in risk over time for a patient \cite{sjolander2024test}. In settings like healthcare, high-dimensional longitudinal information on the patient's state can be an informative source for the prediction of clinical risk of mortality and other events. \\

\noindent To incorporate longitudinal, high-dimensional, and potentially multimodal data and to move away from constraining assumptions of simple statistical models, deep learning methods have been proposed in survival analysis. These include extensions to the Cox model which estimate the parameters using a deep learning model. However, these models remain restrained with the proportionality assumption as Cox, such as the case of DeepSurv \cite{katzman2018deepsurv}. Existing methods often rely on parametric or semi-parametric assumptions of the survival distribution, thereby potentially restraining the predicted distribution away from a closer representation of the true survival distribution \cite{wiegrebe2024deep}. Other deep learning models include DeepHit and its extension to longitudinal data, Dynamic-DeepHit, which implements a custom loss and a simple recurrent neural network to avoid making any parametric assumptions on the survival distribution \cite{lee2018deephit, lee2019dynamic}. Simple extensions of recurrent neural networks for estimating survival distributions have also been proposed \cite{giunchiglia2018rnn, ren2019deep}. More recently, the autoencoder structure has been proposed to learn from a combination of data modalities by finding a lower-dimensional latent representation of the data. ConcatAE uses a simple autoencoder on multi-omics data to find a hidden representation then used in a multi-class classification task to predict discrete survival times \cite{tong2020deep}. This treats the survival analysis task as one of simple classification and does not aim to estimate the survival distribution at all. On the other hand, the variational autoencoder present in VAECox, directly estimates the survival distribution using the Cox loss function while taking advantage of the variational inference over traditional autoencoders \cite{kim2020improved}. Due to the reliance on the Cox loss, VAECox suffers from the often-violated proportionality assumption meaning its results in real-world datasets could be limited. \\

\noindent Our work addresses these limitations by using a cumulative incidence risk estimation loss function based on the negative log-likelihood which requires making no parametric or proportionality assumptions on the survival distribution and hazard risks. We also extend the variational setting to conditional VAE which we find improves predictive performance. Our model, DySurv, is compatible with both static and longitudinal time-series data enabling more comprehensive learning from patient electronic health records. We validate our approach both in static and time-varying settings using six benchmark datasets as well as two large public open-access ICU datasets.The critical care setting provides high-frequency longitudinal measurements and short-term health outcomes in which dynamic risk stratification can help with urgent prognostication and prevention over just detection \cite{dummitt2018using}. We aim to show that conditional variational inference and autoencoder reconstruction tasks can improve learning from complex time-series data by extracting latent features to optimise the survival task. DySurv accomplishes this by adding a conditional variational inference loss to the cumulative risk loss estimation from logistic hazards which aids in learning the survival task. Since DySurv is not only multimodal in input but also in output with both reconstruction and survival tasks included, conditional variational autoencoders (CVAE) provide better predictive performance \cite{itkina2020evidential}. By mapping high-dimensional covariates into a lower-dimensional space, CVAEs can effectively capture the underlying latent structure of the data to estimate survival outcomes. This capability allows deeper insights into dynamic mortality risk prediction in healthcare.\\

\section{METHODS}
\subsection{Data}
\noindent Standard benchmark datasets contain only static features but here we implement survival analysis on the ICU datasets from MIMIC-IV and eICU which contain both static and time-series data. To show performance across different datasets and with different sizes, we will succinctly introduce these datasets. Across all datasets, the event in question is death. The datasets were split into 60\% training, and 20\% each for validation and testing. Quantile transformations have been applied for standardisation and fit only on the training dataset. Please see the Supplementary for a full list of features for each dataset. \\

\subsubsection{SUPPORT}
\noindent The Study to Understand Prognoses and Preferences for Outcomes and Risks of Treatments contains data from five care academic centres in the United States for patient survival in the following six months \cite{knaus1995support}. The result of the study was a prognostic model to estimate survival for seriously ill hospitalised patients. The dataset consists of 8,873 samples and 14 static features. A total of 68.1 percent of patients died with a median death time of 58 days.\\

\subsubsection{METABRIC}
\noindent The Molecular Taxonomy of Breast Cancer International Consortium contains genetic and clinical data from breast cancer patients with 1,904 samples and 9 static features \cite{curtis2012genomic}. 57.7\% have an observed death with a median survival time of 116 months. \\

\subsubsection{GBSG}
\noindent The Rotterdam \& German Breast Cancer Study Group contains treatment and clinical data on 2,232 breast cancer patients with 6 static features \cite{schumacher1994randomized}. 79.6\% of patients have an observed event. \\

\subsubsection{NWTCO}
\noindent The National Wilm's Tumor dataset contains staging and clinical data on 4,028 Wilms' tumour patients with 6 static features \cite{breslow1999design}. 16.0\% of patients have an observed event. \\

\subsubsection{sac3}
\noindent The simulated dataset contains discrete event times with 44 static features and 100,000 samples \cite{kvamme2019continuous}. 37\% of samples have been censored. \\

\subsubsection{sac\_admin5}
\noindent The simulated dataset contains discrete event times with 5 static features and 50,000 samples \cite{kvamme2019brier}. 37\% of samples have been censored.\\

\subsubsection{MIMIC IV and eICU}
We conduct experiments on the de-identified real-world ICU dataset Medical Information Mart for Intensive Care (MIMIC-IV v. 2.0, July 2022) including discharge information for more than 15,000 additional ICU patients compared to the previous release \cite{johnson2020mimic}. The dataset contains data from Beth Israel Deaconess Medical Center collected between 2008 and 2019. The dataset contains 71,935 samples of ICU stays with 33 static features (categorical features were one-hot encoded) and 65 time-varying features. Furthermore, we tested survival analysis models in the eICU Collaboration Research Database \cite{pollard2018eicu}. The eICU database was processed using postgreSQL and the \textit{pandas} package. eICU is a multi-centre ICU database with over 200,859 patient unit encounters for 139,367 unique patients admitted between 2014 and 2015 to one of 335 ICUs at 208 hospitals located throughout the United States. The database is de-identified and includes vital sign measurements, demographic data, and diagnosis information. Static variables include age, sex, admission unit, and others that did not have missingness. For the time-series variables, we use forward filling as clinicians in practice would only consider the last recorded measurement and as has been done by previous work on these datasets \cite{rocheteau2021temporal, nordmark2021practice}. If the first set of measurements is missing for some time-varying features, instead of dropping those features or patients, we backward fill from the closest measurement in the future. 10.0\% of patients had an observed event of death. The pre-processing of MIMIC-IV and eICU follows from previously published work on these datasets but is adapted for the survival scenario with the duration of stay in the ICU or the maximum time horizon for the event times defined as 10 days and time-series features taken in 72-hour timesteps after resampling to 1-hour intervals \cite{mesinovic2023xmi}. 9.5\% of patients had an observed event of death in this cohort. Censoring, as always, is defined as loss-to-follow-up or discharge.

\subsection{DySurv}
\noindent In survival analysis, the main underlying goal is the estimation of the survival function which represents the probability that no event occurs until a time \textit{t} and can be written as
\begin{equation} \label{eq:survival_ftn}
S(t)=\operatorname{P}(T>t)=\int_t^{+\infty} f(u) d u=1-F(t)
\end{equation}
where $f(t)$ is the probability density function of event time and
\begin{equation}
{f(t)=-\frac{dF}{dt}}
\end{equation}
and $F(t)$ corresponds to the cumulative risk or incidence function $F(t)=\operatorname{P}(T \leq t)$. $T$ represents the time of the event, $\operatorname{P}$ is the probability, and $t$ is the specitic timestamp for risk estimation. The key to training a deep learning model is to learn an estimate of the cumulative incidence or risk function $\hat{F}(t)$ as the joint distribution of the event time and outcome label given the observations. In general, the time-to-event values can be left to be continuous depending on the model being considered but we discretise the time set into 10 equally spaced time periods for each dataset in the fashion of DeepHit and Dynamic-DeepHit \cite{lee2019dynamic}. As we discretise the time into intervals, we can estimate this event probability across arbitrary periods and remain faithful to the original survival analysis formulation rather than resorting to chained binary classification \cite{sun2021attention}. Furthermore, measurements are often right-censored, meaning that patients can leave the study or be lost to follow-up and they therefore do not experience an event at all. A common assumption is that this censoring is not important and independent of the outcome of the study itself \cite{tsiatis2004joint}. Once we have an estimate of the risk, we can then simply obtain the survival function and curves by subtracting the cumulative risk from 1 as shown in equation \ref{eq:survival_ftn}. \\

\noindent Data for survival analysis contain three main sets of variables, the first is the feature set which can consist of static or time-series features (the latter having measurements at potentially different sampling frequencies), the time-to-event for the events in question or censoring, respectively, and the outcome label for an event \cite{bradburn2003survival}. Thus, the dataset can be represented as 

\begin{equation}
\mathcal{D}=\left\{\left(\mathcal{X}^i, \mathcal{T}^i, y^i\right)\right\}_{i=1}^N
\end{equation} 
with $\mathcal{X}^i$ representing the feature matrix of patient $i$, $\mathcal{T}$ being the time-to-event i.e., the minimum of the event and censoring time, $y\in\{\varnothing, 1\}$ being the label for the outcome with $\varnothing$ representing right-censoring, and $N$ samples included. Static features have been expanded into time-series by replication padding matching the respective time-series component. Effectively, the feature matrix then consists only of time-series features. A feature matrix for patient $i$, $\mathcal{X}^i$, can be seen as
\begin{equation}
\mathcal{X}^i=\left\{\boldsymbol{x}_1^i, \boldsymbol{x}_2^i, \ldots, \boldsymbol{x}_j^i\right\}
\end{equation}
where $j$ is the length of the time-series for $1 \leq j<J^i$ where $J$ is the maximum time step with timestamps of measurements $\left[t_1^i, t_2^i, \ldots, t_j^i\right]$. $\boldsymbol{x}_j^i$ contains $M$ set of features $\left[x_{j, 1}^i, x_{j, 2}^i, \ldots, x_{j, M}^i\right]^\top$ for timestamp $t_j^i$. \\

\noindent We undertake estimation of the underlying cumulative risk by mainly optimizing the negative logarithmic likelihood of the joint distribution of the event time and outcome with right-censoring. For those patients who have suffered the event, we capture both the outcome and the time at which it occurs. For censored patients, we capture the censoring time conditioned on the measurements recorded prior to the censoring. If we assume $\hat{a}_{t}=P\left(T=t \mid \mathcal{X}\right)$ represents the estimated probability of experiencing an event at time $t$, then the loss can be represented as

\begin{equation} \label{eq:CIF}
\begin{aligned}
\mathcal{L}_1=- & \sum_{i=1}^N\left[\mathbb{1}\left(y^i \neq \varnothing\right) \cdot \log \left(\frac{\hat{a}_{t^i}^i}{1-\sum_{y \neq \varnothing} \sum_{n \leq t_{J i}^i} \hat{a}_{n}^i}\right)\right. \\
& \left.+\mathbb{1}\left(y^i=\varnothing\right) \cdot \log \left(1-\sum_{y \neq \varnothing} \hat{F}\left(t^i \mid \mathcal{X}^i\right)\right)\right]
\end{aligned}
\end{equation}
\noindent where $\mathbb{1}$ is the indicator function. The first term represents optimising for cumulative incidence risk accounting for uncensored patients and the second for those censored at the last noted time ie. alive. The estimated cumulative incidence risk, $\hat{F}$, is the probability that the patient dies at time $t$ or before, conditioned on all previous longitudinal measurements. By optimising for this loss, we estimate the actual risk distribution for each patient and a prediction can be made for arbitrary times across event times. \\

\noindent As data are complex (i.e., static and time-varying), we seek to learn a latent representation using a conditional variational autoencoder that would improve learning for the task of survival analysis \cite{zhang2016variational}\footnote{For those readers less familiar with the autoencoder architecture and machine learning background for survival analysis, we recommend the following insightful surveys and chapters to peruse: \cite{wiegrebe2024deep, wang2019machine, pinheiro2021variational}}. Since we are working with a deep learning model, our goal is to minimise a specific loss function relevant to our task. The loss will penalise our model when it starts learning away from the task in question. The first loss term relates to learning an efficient cumulative risk distribution to represent survival as described above in Equation \ref{eq:CIF}. There are two further loss terms, namely the reconstruction (mean squared error) and the Kullback–Leibler (KL) divergence in the variational autoencoder \cite{fil2021beta}. The first helps the model learn better representations of the input data, while the second helps learn better latent representations. Estimating the distribution of the underlying latent factors relies on minimising the KL divergence between an approximation of the true posterior and the true distribution both of which are assumed to be multivariate Gaussians. As such, learning these distributions means optimising for their parameters, the mean and standard deviation. $z$ is the sampled latent vector from the probabilistic encoder for the learned Gaussian distribution with mean $\mu$ and standard deviation $\sigma$. For training, however, since sampling is a stochastic process, we use the reparameterization trick to backpropagate the gradient and represent the latent vector as the sum of a deterministic variable and an auxiliary independent random variable $\varepsilon$ \cite{kingma2013auto}

\begin{equation}
\varepsilon \sim N(0,1) \quad z=\mu+\varepsilon * \sigma \quad \rightarrow \quad z \sim N(\mu, \sigma)
\end{equation}

\noindent The ability of the decoder to successfully reconstruct the input is captured with a simple mean squared error term between the reconstruction of the input and the input itself. Thus, the loss for variational inference can be seen as

\begin{equation}
\begin{aligned}
\mathcal{L}_2=L_{v a e}(E, D)=\left\|\mathcal{X}-\mathcal{X}_{\text {recon }}\right\|_2 \\
+\frac{1}{2} \sum_{i=1}^{z_{\text {dim }}}[\left(\mu_i^2+\sigma_i^2\right)-1-\log \left(\sigma_i^2\right))]
\end{aligned}
\end{equation}

\noindent The latent vector $z$ is then used as input to a neural network module in optimising the survival task. During training, the decoder uses the latent vector and the condition vector (survival labels or times of death) as input which helps the latent space to capture other information instead of trying to better reconstruct the input. At test time, the decoder is not used anymore, and the latent space is used for prediction in the survival task. The total loss can then be presented as
\begin{equation}
\mathcal{L}=\alpha \mathcal{L}_1+(1-\alpha) \mathcal{L}_2
\end{equation}
where $\alpha$ is the balancing coefficient between the two losses ($\mathcal{L}_1$ and $\mathcal{L}_2$) and $0 \leq \alpha \leq 1$. $\alpha$ is considered as a hyperparameter that is optimised during training according to multi-objective optimization principles. \\

\begin{figure}[H]
    \centering
    \includegraphics[width=0.9\linewidth]{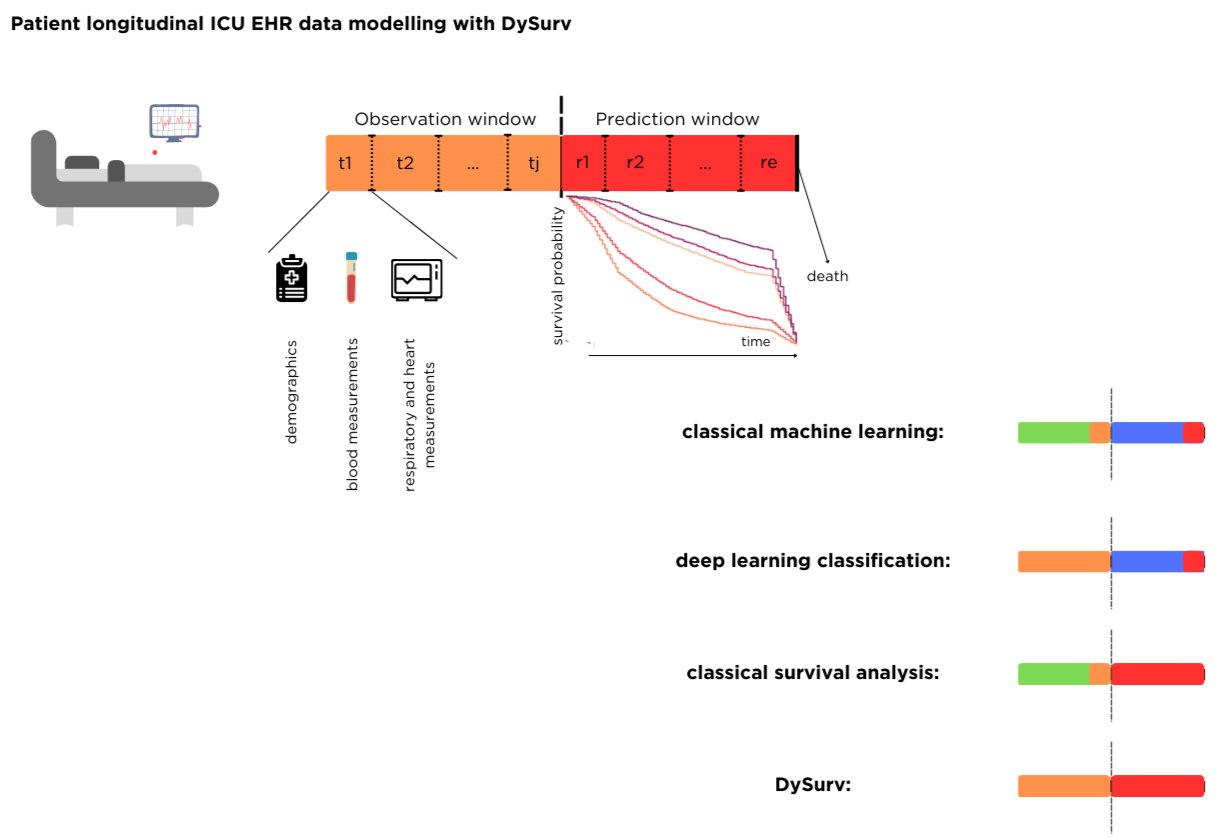}
    \caption{(a) Description of the proposed DySurv framework using longitudinal EHR data for dynamic risk prediction instead of fixed-point event classification. The patient stay consists of measurements until $t_{j}$ as the last measurement recorded in the observation window. The orange marker states how much of the longitudinal data, i.e., how many timestamps are used in the model learning process. The red marker indicates how many timestamps into the future the model estimates the risk of death. In the first instance of classical machine learning, only the most recent timestamp measurement is used to predict the risk of death at a fixed time point in the future at a prediction window distance (except for Gaussian processes). For deep learning classification, by using LSTMs, we can learn from the entire patient longitudinal measurement but only estimate the risk of death at a fixed timepoint, hence the observation window is fully orange, but only one of the prediction window timestamps is red. DySurv uses all of the available longitudinal data to predict risk dynamically, ie. at all reasonable times into the feature.}
    \label{fig:DySurv2}
\end{figure}

\noindent DySurv leverages the dynamic nature of deep learning time-series models combined with conditional variational autoencoders for multi-task learning of survival analysis risk prediction extending beyond the classical fixed binary event prediction from traditional machine learning models. Figure \ref{fig:DySurv2} shows the framework employed here built around DySurv to support robust risk estimation for clinical events such as death. Using a simple autoencoder has been shown to lead to overfitting and imbalanced learning of the reconstruction task that could harm learning the survival task whereas a VAE's objective function is based on the reconstruction loss from a randomly sampled vector allowing for more robustness \cite{kim2020improved}. We concatenate the feature vectors from the static and time-series features together before feeding them into an encoder equipped with a Long-short-term-memory (LSTM) cell. The LSTM module allows us to learn the temporal patterns in the longitudinal data and represent it in a hidden embedding for downstream tasks such as the estimation of the survival function. A short primer on the LSTM cell can be found in Supplementary Section C. \\

\noindent The time-series components are compressed into a latent representation that is then used, as is, as input to the estimation of the negative log-likelihood loss function. A detailed description of the architecture can be found in the Supplementary. Once the latent vector is sampled from the Gaussian distribution defined by these parameters, the lower-dimensional latent factors are used as input for an MLP module to optimize the survival task. In the case of the ICU datasets, for example, each output node of this MLP, a representation of $\hat{a}_{t}$, then corresponds to a 1-day risk prediction in the ICU for a maximum time horizon of 10 days. The hyperparameters optimised in the network through grid search using the training and validation set included learning rate, batch size, $\alpha$, and dropout proportion. To minimise overfitting, we employ early stopping.\\

We compare the performance of our model to a collection of survival analysis models summarised in Table I and described in detail in the Supplementary. Unlike DySurv, existing methods like Cox-based methods as well as some deep learning alternatives, rely on assuming a specific statistical distribution of the survival distribution, thus making parametric distributions that restrict their predictive power. This assumption is described in further detail in Supplementary Section B.\\

\begin{table}
\caption{\textbf{Survival analysis methods investigated}}
\small
\centering
\begin{tabular}{@{}lll@{}}
\toprule
    \textbf{Method} & \textbf{Time scale} & \textbf{Reference}\\
    \midrule
PMF & discrete-time & \cite{kvamme2019continuous}\\
\addlinespace[0.05cm]
MTLR & discrete-time & \cite{yu2011learning}\\
\addlinespace[0.05cm]
BCESurv & discrete-time & \cite{kvamme2019brier}\\
\addlinespace[0.05cm]
DeepHit & discrete-time & \cite{lee2018deephit}\\
\addlinespace[0.05cm]
Logistic Hazard & discrete-time & \cite{gensheimer2019scalable}\\
\addlinespace[0.05cm]
CoxTime & continuous-time & \cite{kvamme2019time}\\
\addlinespace[0.05cm]
CoxCC & continuous-time & \cite{kvamme2019time}\\
\addlinespace[0.05cm]
DeepSurv & continuous-time & \cite{katzman2018deepsurv}\\
\addlinespace[0.05cm]
PCHazard & continuous-time & \cite{kvamme2019continuous}\\
\bottomrule                          
\end{tabular}
\label{tab:models}
\end{table}

\subsection{Metrics}
\noindent In this section, we will switch the notation of samples from superscript to subscript, hence $\mathbf{x}^i$ is now $\mathbf{x}_i$ for sample $i$. Since we are no longer making single risk predictions at specific times and estimating distributions of event times for censored samples, different evaluation metrics must apply than those used in classic machine learning classification and prediction settings. The most common metric for evaluating survival analysis models is the \textit{concordance index $C_{ind}$}, which estimates the probability that, for a random pair of samples, the predicted survival times (risk probabilities) of the two samples have the same ordering as their true survival times \cite{kvamme2019time}. This explanation works perfectly for settings of proportional hazards where the ordering does not change over time but for our purposes, we will not be limited by such assumptions. Hence, we will rely on using the time-dependent extension $C_{ind}^{\mathrm{td}}$ with some modifications accounting for predictions independent of feature observations having a concordance of 0.5. The metric was originally proposed in \cite{antolini2005time} and all of the included metrics are provided in the PyCox package. The metric can be represented as

\begin{equation}
C_{ind}^{\mathrm{td}}=\mathrm{P}\left\{\hat{S}\left(T^i \mid \mathbf{x}^i\right)<\hat{S}\left(T^i \mid \mathbf{x}^j\right) \mid T^i<T^j, y^i=1\right\}
\end{equation}

\noindent where $\hat{S}$ indicates the estimated survival probabilities are used and $y^i=1$ that only those who experienced the event are considered in this metric. A noted limitation of this metric is its obvious bias and dependence on the censoring distribution as only non-censored samples are considered making it affected by the length of stay and the censoring proportion that increases over the length of stay. To this end, we decided to use additional metrics for more holistic evaluation especially as previously proposed models like DeepHit were found to have lackluster results in real-world datasets when evaluated using other metrics besides concordance. We also evaluate our model using the \textit{Integrated Brier Score} or \textit{IBS}. The Brier Score is similar to the mean squared error as it represents the average squared distances between the predicted and the true survival probability (approximated with step functions with jumps at the event times) and is always a number between 0 and 1, with 0 being the best possible value \cite{fotso2018deep}. The expectation of the Brier Score contains the mean squared error as one of its additive terms, so minimising one is minimising the other \cite{kvamme2019brier}. Since we need to know the event times for calculating IBS and we do not have access to all the samples' event times in right-censoring, an adjusted metric called the inverse probability of censoring weights Brier Score (IPCW) is used instead to approximate the times by weighting the scores of the observed event times by the inverse probability of censoring. The equation used is thus 

\begin{equation}
\begin{aligned}
\operatorname{BS}_{\operatorname{IPCW}}(t)=\frac{1}{n} \sum_{i=1}^n[\frac{\hat{S}^i(t)^2 \mathbb{1}\left\{T^i \leq t, y^i=1\right\}}{\hat{G}^i\left(T^i-\right)}+ \\
\frac{\left[1-\hat{S}^i(t)\right]^2 \mathbb{1}\left\{T^i>t\right\}}{\hat{G}^i(t)}]
\end{aligned}
\end{equation}

\noindent where $\hat{G}^i(t)=\mathrm{P}\left(C^{i*}>t\right)>0$ is the Kaplan-Meier estimate of the censoring distribution for sample $i$ and $C^{i*}$ is the censoring time. The expected value of this metric is the same as that for the uncensored Brier Score. As one notices, this metric is evaluated at specific times, whereas the \textit{ Integrated Brier Score} or \textit{IBS} provides a general evaluation of model performance at all times.

\begin{equation}
\mathrm{IBS}=\frac{1}{\max \left(T^i\right)} \int_0^{\max \left(T^i\right)} BS_{\operatorname{IPCW}}(t) d t
\end{equation}

\noindent A limitation of this metric, however, is the biased assumption of the censoring distribution being the same across samples thereby disregarding covariate effects. This can be addressed by using an administrative extension of the metric that requires access to all the censoring times but a discussion of this in greater detail can be perused where the metric was originally proposed \cite{kvamme2019brier}. Lastly, we introduce the \textit{IPCW (negative) binomial log-likelihood} or \textit{NBLL} from classic binary classification from \cite{kvamme2019brier} which measures both discrimination and calibration of the estimates and uses its integrated extension, \textit{INBLL}, for all times

\begin{equation}
\begin{aligned}
\operatorname{BLL}(t)=\frac{1}{N} \sum_{i=1}^N[\frac{\log \left[1-\hat{S}\left(t \mid \mathbf{x}^i\right)\right] 1\left\{T^i \leq t, y^i=1\right\}}{\hat{G}\left(T^i\right)}+ \\
\frac{\log \left[\hat{S}\left(t \mid \mathbf{x}^i\right)\right] 1\left\{T^i>t\right\}}{\hat{G}(t)}]
\end{aligned}
\end{equation}

\begin{equation}
\mathrm{IBLL}=\frac{1}{\max \left(T^i\right)} \int_{0}^{\max \left(T^i\right)} \operatorname{BLL}(t) d t
\end{equation}

\noindent For both of the last metrics, we approximate the integrals by numerical integration (for 100 timesteps as based on previous literature), and the time span is the duration of the test set as these metrics are only evaluated on the test set \cite{kvamme2019time}. 

\section{RESULTS}
\noindent To holistically evaluate DySurv, we present a set of experiments and comparisons with other benchmark survival analysis models across multiple datasets. We present not only the discriminative performance of the model as measured by concordance but also its calibration as measured by IBS and IBLL. We evaluated the model on a case study example of patients in the ICU by including the real-world MIMIC-IV and eICU electronic health record datasets. The results consist of two major experiments, one is the ability of the model to successfully learn from static data which is present in all the datasets, and the other to learn from a combination of static and time-varying data such as in MIMIC-IV and eICU. For these purposes, Dynamic-DeepHit is the only relevant comparison as other survival analysis models deal only with static data. Tables \ref{tab:Result}, \ref{tab:Result2}, and \ref{tab:Result_eICU} show these results across datasets on held-out test sets for all models included. We implemented these methods in PyTorch (PyCox) v. 1.10.1 using a MacBook Air M1 2021 laptop with data processing completed using pandas and SQL. \\

\begin{table*}[t]
    \centering
    \caption{Test results on different static-only datasets for survival analysis models and DySurv as evaluated by three different metrics introduced in Materials and Methods. For concordance, higher is better, and for the other two metrics, lower is better. The best results are in bold. All of the results are an average of five random seeds.}
    \label{tab:Result}
    \setlength{\tabcolsep}{10pt}
    \begin{tabularx}{0.9\textwidth}{l
                                    c
                                    c
                                    c
                                    l
                                    c
                                    c
                                    c
                                 }
    
        \toprule
         & $C_{ind}^{\mathrm{td}}$  & IBS  & IBLL & &$C_{ind}^{\mathrm{td}}$  & IBS  & IBLL\\
         \midrule
         \midrule
         \textit{SUPPORT} &&&& \textit{METABRIC} &&&\\
          \midrule
          \midrule
          PMF & 57.9 & 0.195  & 0.574 & PMF & 63.8 & 0.168 & 0.497\\
          \midrule
          MTLR & 55.3 & 0.205 & 0.775 &MTLR & 56.8 & 0.172 & 0.527\\
          \midrule
          BCESurv & 55.3 & 0.290 & 2.08 &BCESurv & 56.8 & 0.138 & 0.477\\
          \midrule
          DeepHit & 57.3 & 0.273 & 0.678 &DeepHit & 65.5 & 0.123 & 0.415\\
          \midrule
          Logistic Hazard & 53.5 & 0.206 & 0.762 &Logistic Hazard & 59.0 & 0.163 & 0.498\\
          \midrule
          CoxTime & 59.5 & 0.193 & 0.565 &CoxTime & 65.4 & 0.114 & 0.361\\
          \midrule
          CoxCC & 59.7 & 0.192 & 0.563 & \textbf{CoxCC} & \textbf{65.9} & \textbf{0.166} & \textbf{0.508}\\
          \midrule
          DeepSurv & 60.6 & 0.190 & 0.559 &DeepSurv & 62.4 & 0.176 & 0.541\\
          \midrule
          PCHazard & 55.1 & 0.206 & 0.633 &PCHazard & 51.4 & 0.160 & 0.547\\
          \midrule
          \textbf{DySurv} & \textbf{64.7} & \textbf{0.190} & \textbf{0.561} &DySurv & 64.5 & 0.120 & 0.387\\
          \midrule
          \midrule
          \textit{GBSG} &&&& \textit{NWTCO} &&&\\
          \midrule
          \midrule
          PMF & 68.5 & 0.179 & 0.528 &PMF & 69.7 & 0.122 & 0.389\\
          \midrule
          MTLR & 65.6 & 0.180 & 0.542 &MTLR & 66.8 & 0.109 & 0.403\\
          \midrule
          BCESurv & 65.6 & 0.156 & 0.481 &BCESurv & 69.1 & 0.108 & 0.393\\
          \midrule
          DeepHit & 68.1 & 0.174 & 0.514 &\textbf{DeepHit} & \textbf{71.1} & \textbf{0.118} & \textbf{0.348}\\
          \midrule
          Logistic Hazard & 67.4 & 0.179 & 0.537 &Logistic Hazard & 66.5 & 0.108 & 0.396\\
          \midrule
          CoxTime & 68.4 & 0.171 & 0.510 &CoxTime & 70.7 & 0.110 & 0.343\\
          \midrule
          CoxCC & 59.6 & 0.205 & 0.597 &CoxCC & 70.3 & 0.110 & 0.373\\
          \midrule
          DeepSurv & 68.5 & 0.180 & 0.531 &DeepSurv & 68.3 & 0.115 & 0.391\\
          \midrule
          PCHazard & 55.8 & 0.182 & 0.574 &PCHazard & 60.2 & 0.118 & 0.465\\
          \midrule
          \textbf{DySurv} & \textbf{70.4} & \textbf{0.164} & \textbf{0.499} &DySurv & 70.3 & 0.111 & 0.347\\
          \midrule
          \midrule
        \textit{sac3} &&& &\textit{sac\_admin5}&&&\\
          \midrule
          \midrule
          PMF & 74.3 & 0.125 & 0.391 & PMF & 71.5 & 0.124 & 0.387\\
          \midrule
          MTLR & 65.0 & 0.124 & 0.539 & MTLR & 65.7 & 0.122 & 0.520\\
          \midrule
          BCESurv & 67.8 & 0.163 & 0.586 & BCESurv & 68.4 & 0.164 & 0.505\\
          \midrule
          DeepHit & 74.2 & 0.184 & 0.527 & DeepHit & 71.6 & 0.186 & 0.396\\
          \midrule
          Logistic Hazard & 72.0 & 0.120 & 0.492 & Logistic Hazard & 70.7 & 0.118 & 0.481\\
          \midrule
          CoxTime & 78.7 & 0.117 & 0.362 & CoxTime & 78.5 & 0.117 & 0.362\\
          \midrule
          CoxCC & 76.4 & 0.124 & 0.384 & CoxCC & 76.7 & 0.122 & 0.381\\
          \midrule
          DeepSurv & 76.1 & 0.126 & 0.390 & DeepSurv & 77.4 & 0.119 & 0.371\\
          \midrule
          PCHazard & 64.0 & 0.135 & 0.514 & PCHazard & 65.1 & 0.123 & 0.503\\
          \midrule
          \textbf{DySurv} & \textbf{80.6} & \textbf{0.112} & \textbf{0.359} & \textbf{DySurv} & \textbf{79.6} & \textbf{0.116} & \textbf{0.361}\\
            \bottomrule
    \end{tabularx}
\end{table*}

\begin{table*}[t]
    \centering
    \caption{Test results on MIMIC-IV dataset for survival analysis models evaluated by three different metrics. For concordance, higher is better, and for the other two metrics, lower is better. All of the results are an average of five random seeds.}
    \label{tab:Result2}
    \setlength{\tabcolsep}{15pt}
    \begin{tabularx}{0.6\textwidth}{l
                                    c
                                    c
                                    c
                                 }
    
        \toprule
         & $C_{ind}^{\mathrm{td}}$  & IBS  & IBLL\\
         \midrule
          PMF & 50.9 & 0.126 & 0.389\\
          \midrule
          MTLR & 52.4 & 0.126 & 0.389\\
          \midrule
          BCESurv & 52.2 & 0.157 & 0.473\\
          \midrule
          DeepHit & 54.4 & 0.137 & 0.421\\
          \midrule
          Logistic Hazard & 52.6 & 0.122 & 0.396\\
          \midrule
          CoxTime & 53.1 & 0.122 & 0.337\\
          \midrule
          CoxCC & 52.9 & 0.123 & 0.393\\
          \midrule
          DeepSurv & 54.2 & 0.128 & 0.403\\
          \midrule
          PCHazard & 51.0 & 0.122 & 0.378\\
          \midrule
          DySurv (static) & 55.7 & \textbf{0.111} & 0.360\\
          \midrule
          Dynamic-DeepHit & 56.0 & 0.143 & 0.376\\
          \midrule
          DySurv ($+$ time-series) & \textbf{57.9} & 0.122 & \textbf{0.320} \\
            \bottomrule
    \end{tabularx}
\end{table*}

\begin{table*}[t]
    \centering
    \caption{Test results on eICU dataset for survival analysis models evaluated by three different metrics. For concordance, higher is better, and for the other two metrics, lower is better. All of the results are an average of five random seeds.}
    \label{tab:Result_eICU}
    \setlength{\tabcolsep}{15pt}
    \begin{tabularx}{0.6\textwidth}{l
                                    c
                                    c
                                    c
                                 }
    
        \toprule
         & $C_{ind}^{\mathrm{td}}$  & IBS  & IBLL\\
         \midrule
          PMF & 53.4 & 0.115 & 0.323\\
          \midrule
          MTLR & 54.2 & 0.116 & 0.344\\
          \midrule
          BCESurv & 53.1 & 0.138 & 0.416\\
          \midrule
          DeepHit & 57.9 & 0.126 & 0.332\\
          \midrule
          Logistic Hazard & 54.7 & 0.111 & 0.358\\
          \midrule
          CoxTime & 55.7 & 0.112 & 0.310\\
          \midrule
          CoxCC & 55.1 & 0.109 & 0.358\\
          \midrule
          DeepSurv & 54.2 & 0.128 & 0.403\\
          \midrule
          PCHazard & 53.1 & 0.103 & 0.346\\
          \midrule
          DySurv (static) & 58.1 & 0.103 & 0.322\\
          \midrule
          Dynamic-DeepHit & 58.8 & 0.131 & 0.352\\
          \midrule
          DySurv ($+$ time-series) & \textbf{60.3} & \textbf{0.102} & \textbf{0.319} \\
            \bottomrule
    \end{tabularx}
\end{table*}

\begin{figure}[H]
        \centering
        \begin{subfigure}[b]{0.4\textwidth}   
            \centering 
            \includegraphics[width=\textwidth]{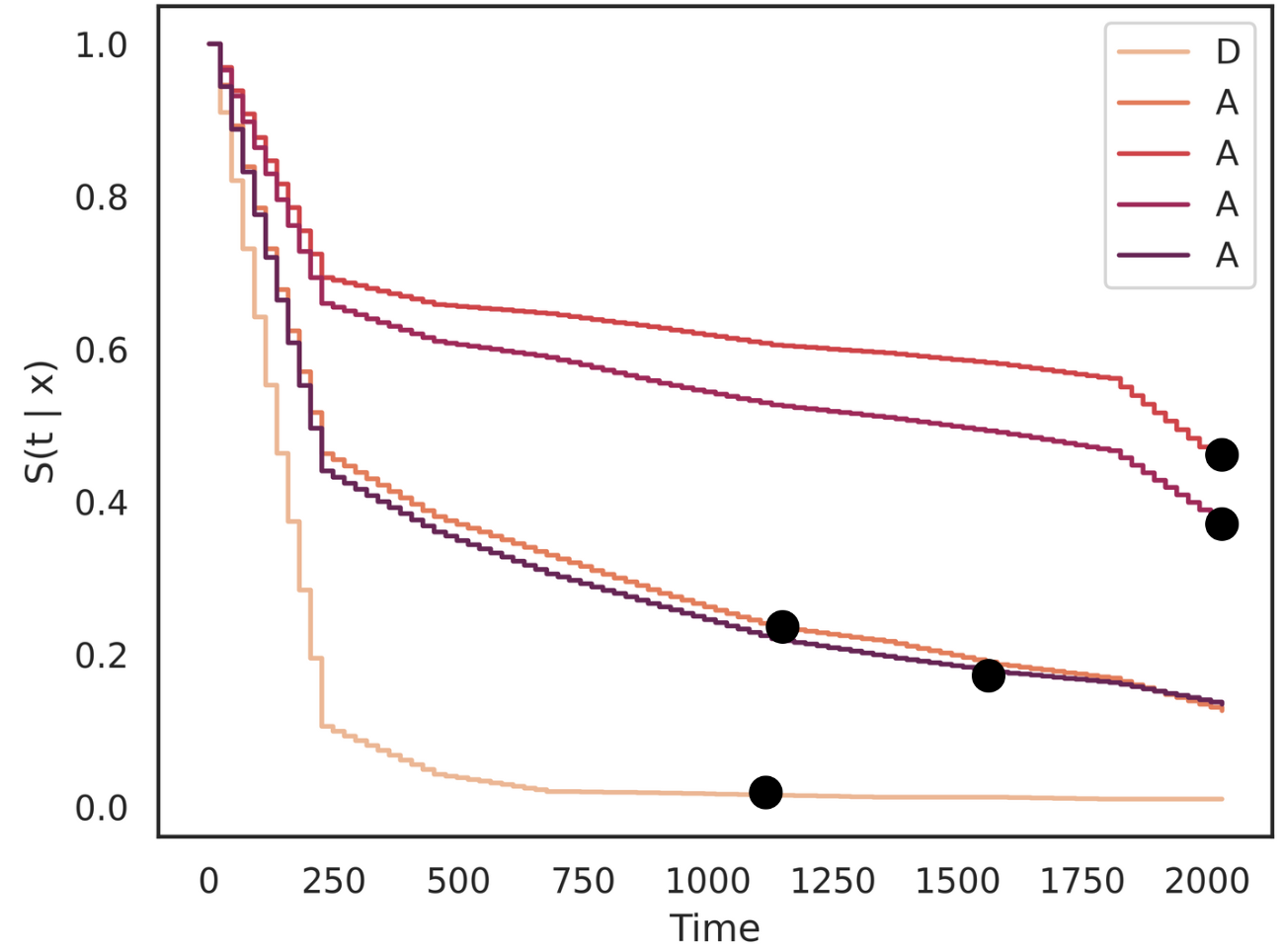}
            \caption[]%
            {{\small Survival curve for five random samples from SUPPORT}}    
            \label{fig:mean and std of net34}
        \end{subfigure}
        \hfill
        \begin{subfigure}[b]{0.4\textwidth}   
            \centering 
            \includegraphics[width=\textwidth]{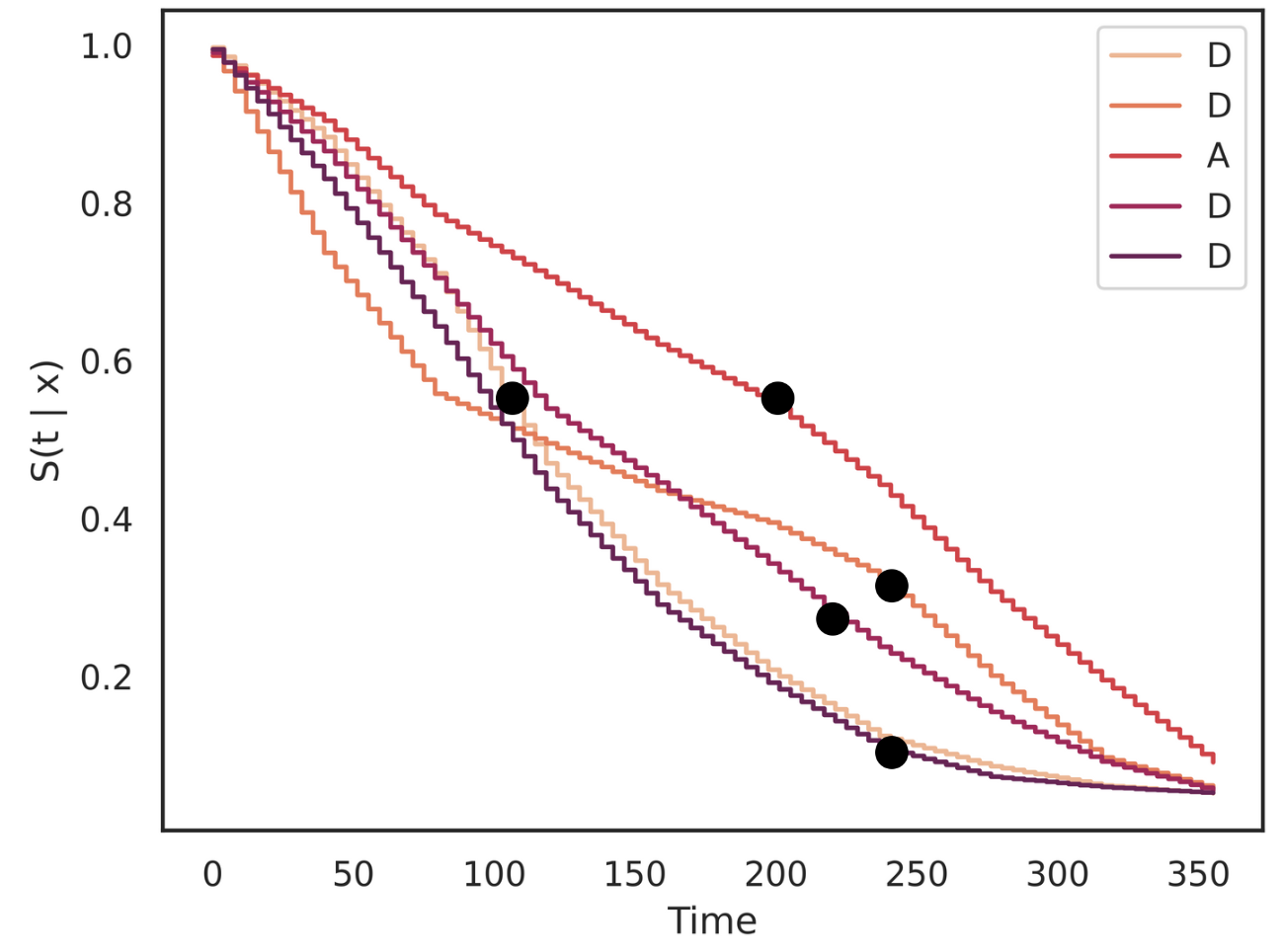}
            \caption[]%
            {{\small Survival curve for five random samples from METABRIC}}
            \label{fig:mean and std of net44}
        \end{subfigure}
        \vfill
        \begin{subfigure}[b]{0.4\textwidth}   
            \centering 
            \includegraphics[width=\textwidth]{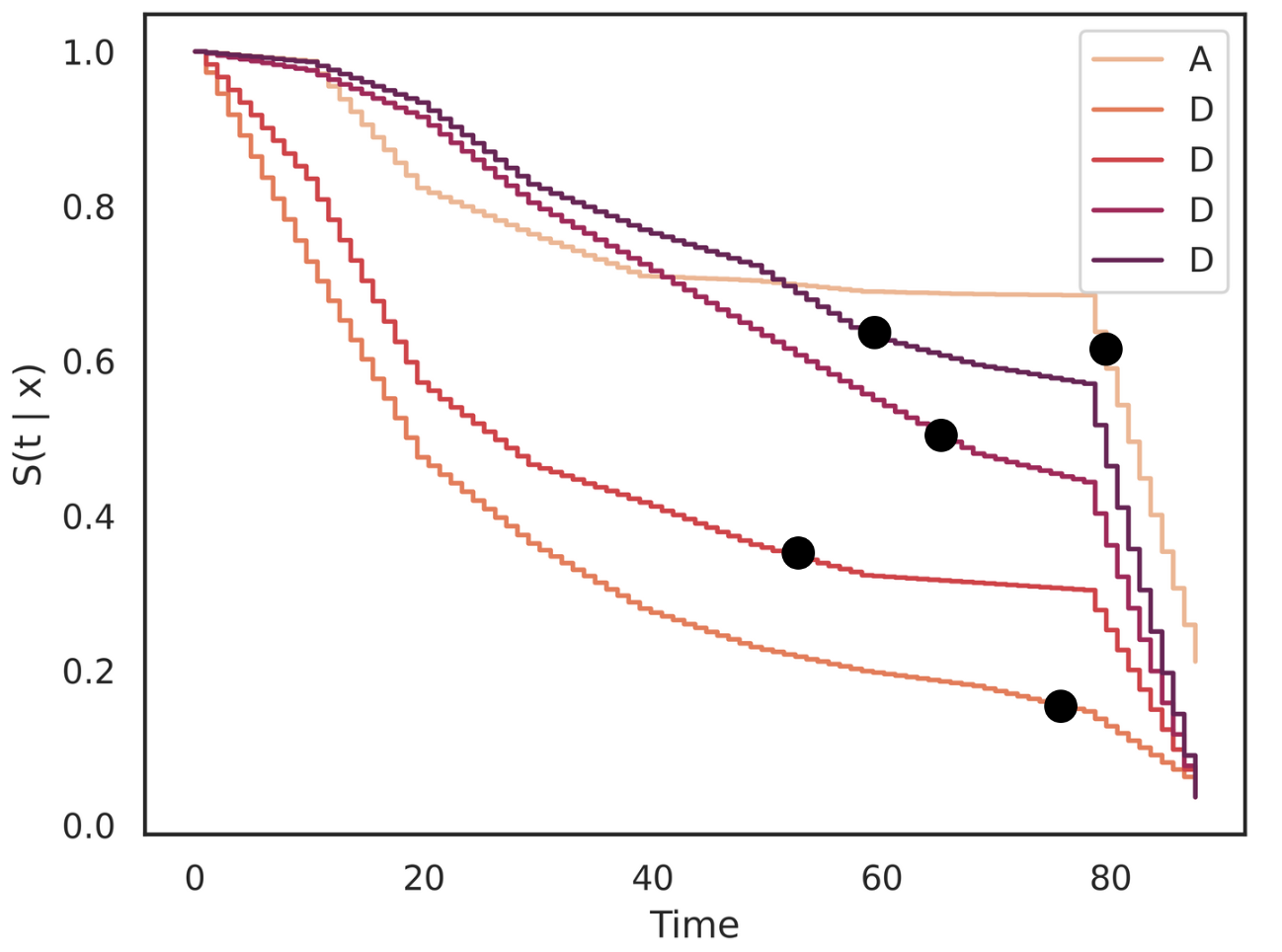}
            \caption[]%
            {{\small Survival curve for five random samples from GBSG}}    
            \label{fig:mean and std of net34}
        \end{subfigure}
        \hfill
        \begin{subfigure}[b]{0.4\textwidth}   
            \centering 
            \includegraphics[width=\textwidth]{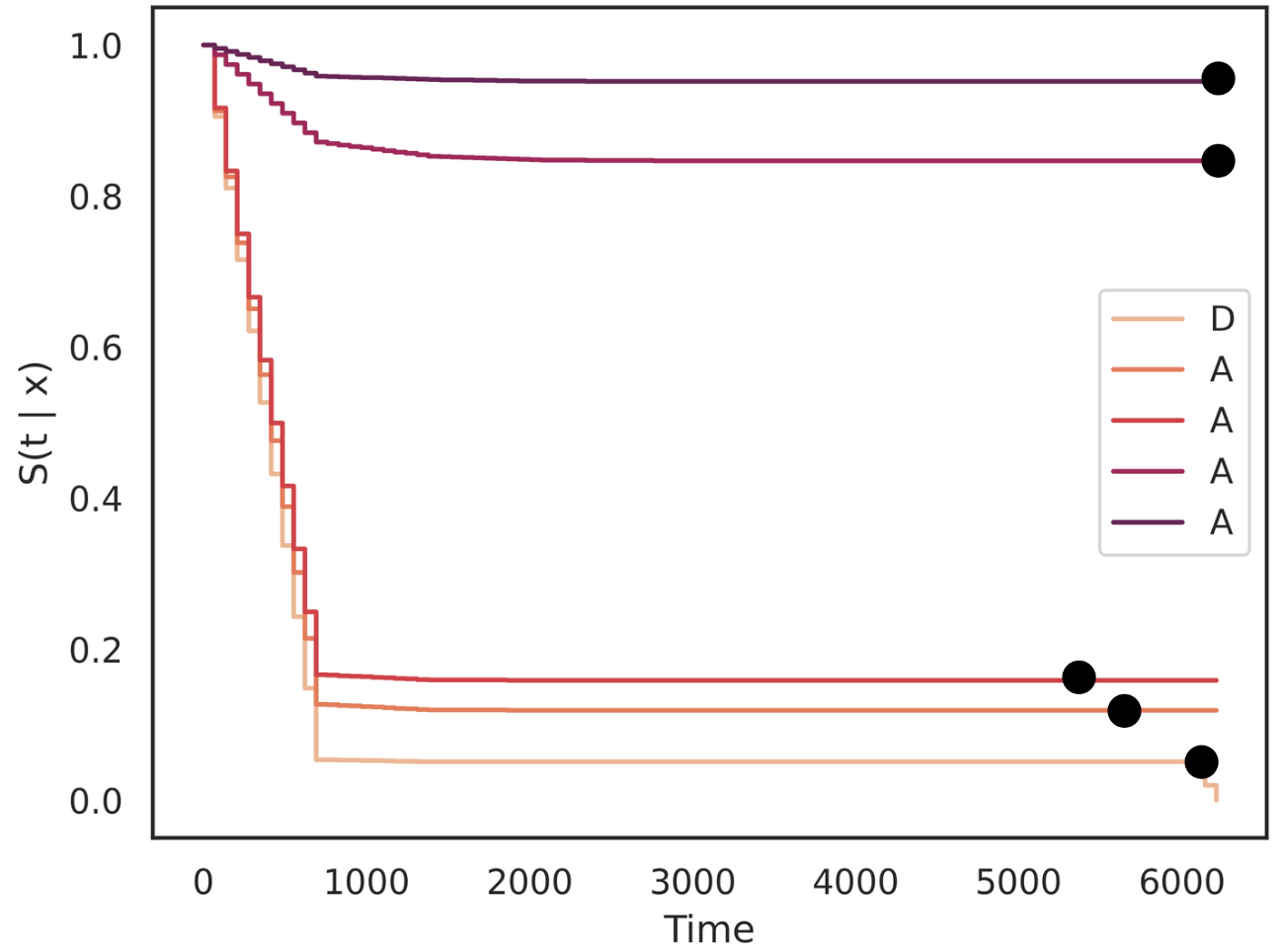}
            \caption[]%
            {{\small Survival curve for five random samples from NWTCO}}
            \label{fig:mean and std of net44}
        \end{subfigure}
        \vfill
        \begin{subfigure}[b]{0.4\textwidth}   
            \centering 
            \includegraphics[width=\textwidth]{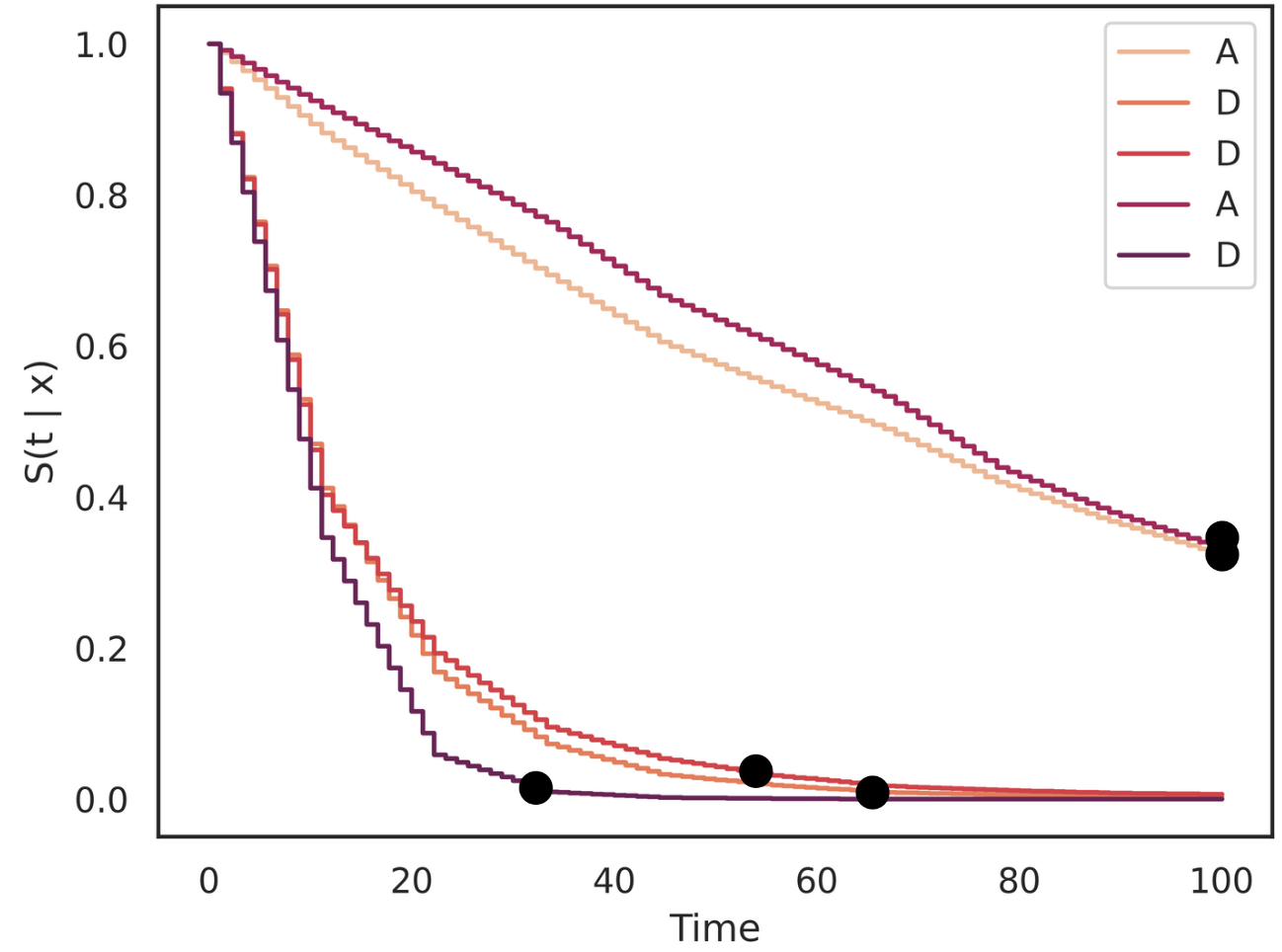}
            \caption[]%
            {{\small Survival curve for five random samples from sac3}}    
            \label{fig:mean and std of net34}
        \end{subfigure}
        \hfill
        \begin{subfigure}[b]{0.4\textwidth}   
            \centering 
            \includegraphics[width=\textwidth]{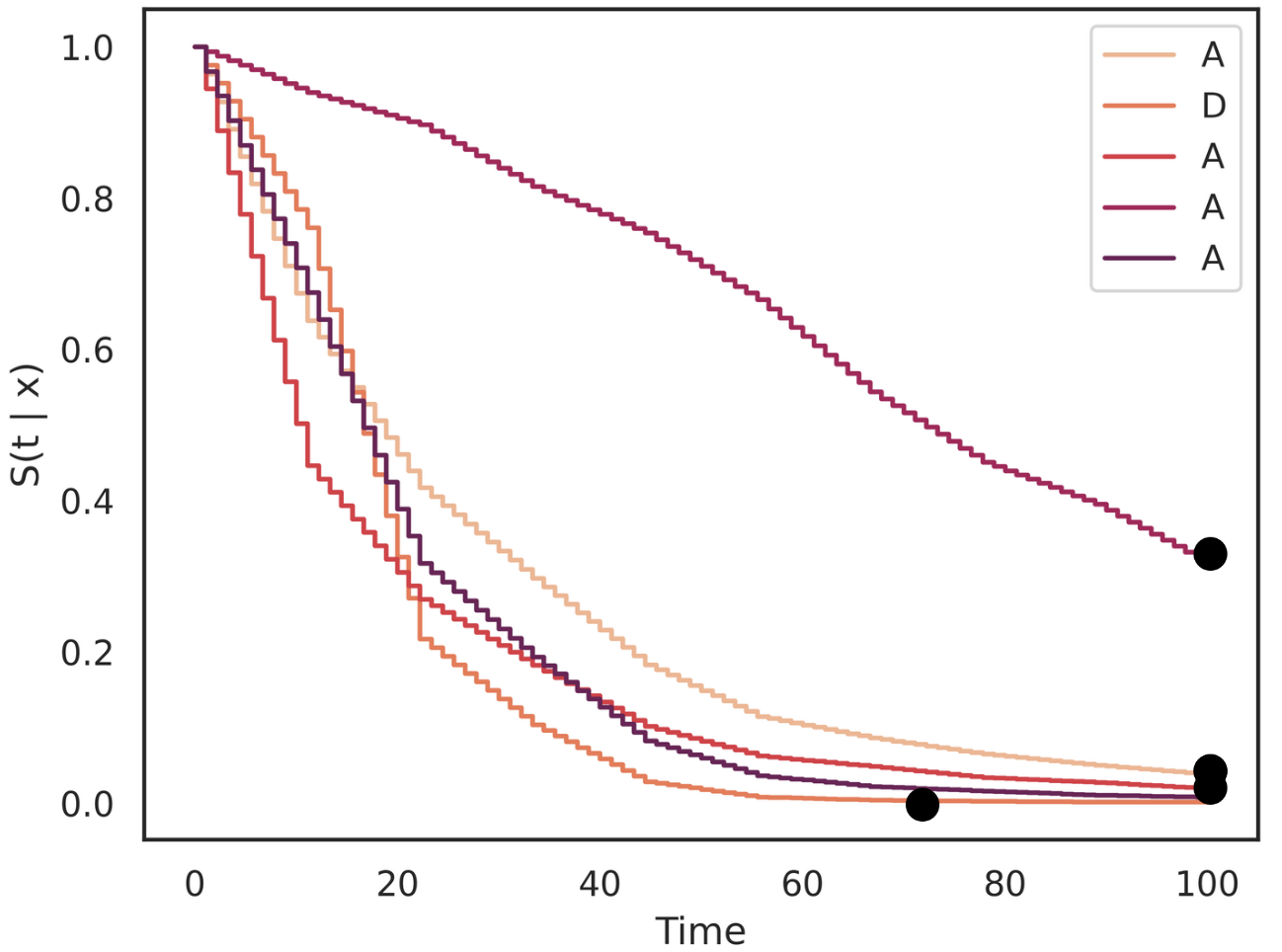}
            \caption[]%
            {{\small Survival curve for five random samples from sac5}}
            \label{fig:mean and std of net44}
        \end{subfigure}
        \caption
        {\small Survival curves (estimate of survival probability over time) for benchmark datasets by DySurv across different samples. DySurv provides discrete estimates over time and additional interpolation was applied. Dots correspond to true event times which were predicted correctly by DySurv.}
        \label{fig:survival_curves}
    \end{figure}

\begin{figure}[H]
        \centering
        \begin{subfigure}[b]{0.46\textwidth}   
            \centering 
            \includegraphics[width=\textwidth]{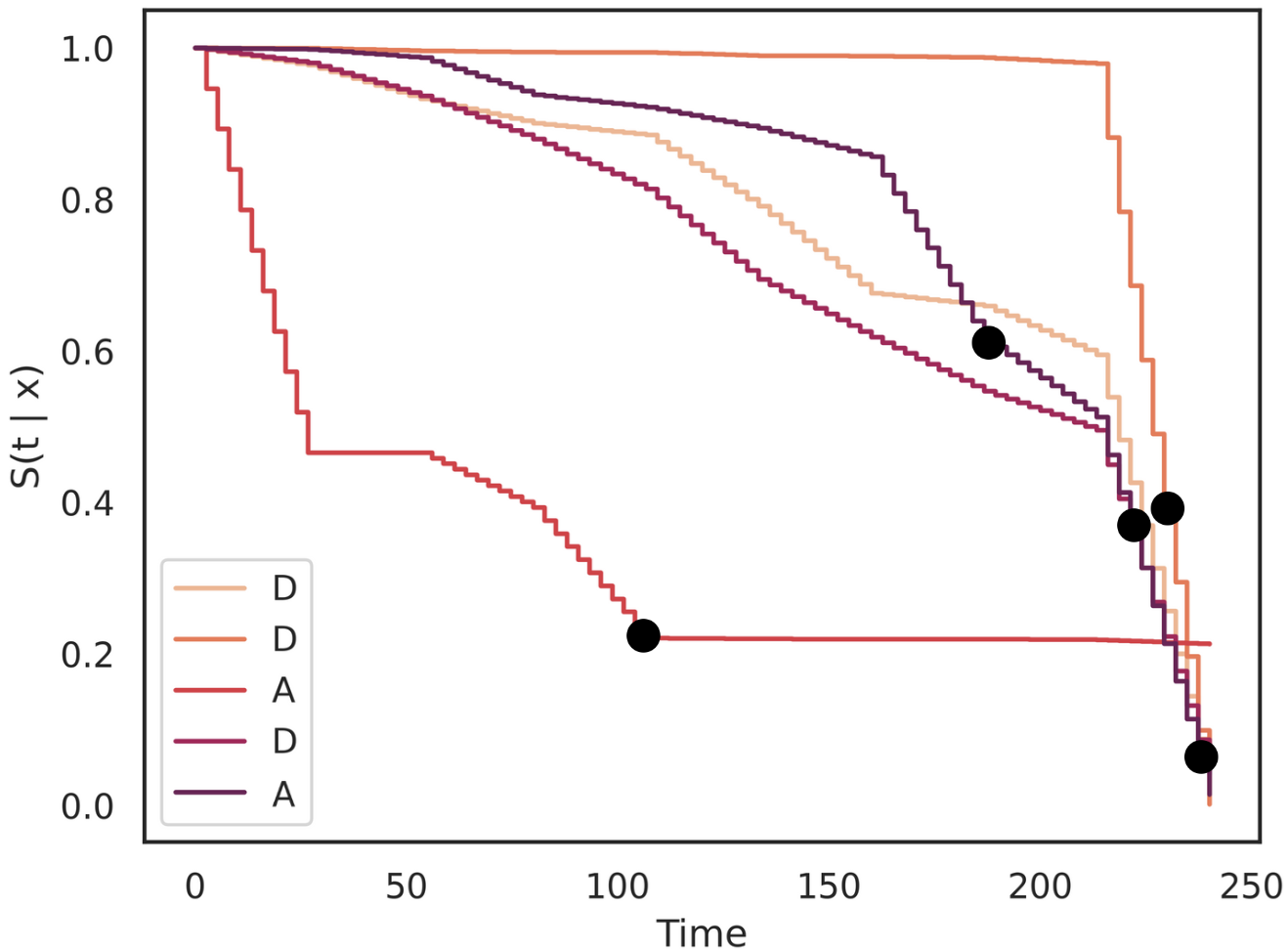}
            \caption[]%
            {{\small Survival curve (estimate of survival probability over time) for five random samples from MIMIC-IV using only static features with DySurv}}    
            \label{fig:mean and std of net34}
        \end{subfigure}
        \hfill
        \begin{subfigure}[b]{0.46\textwidth}   
            \centering 
            \includegraphics[width=\textwidth]{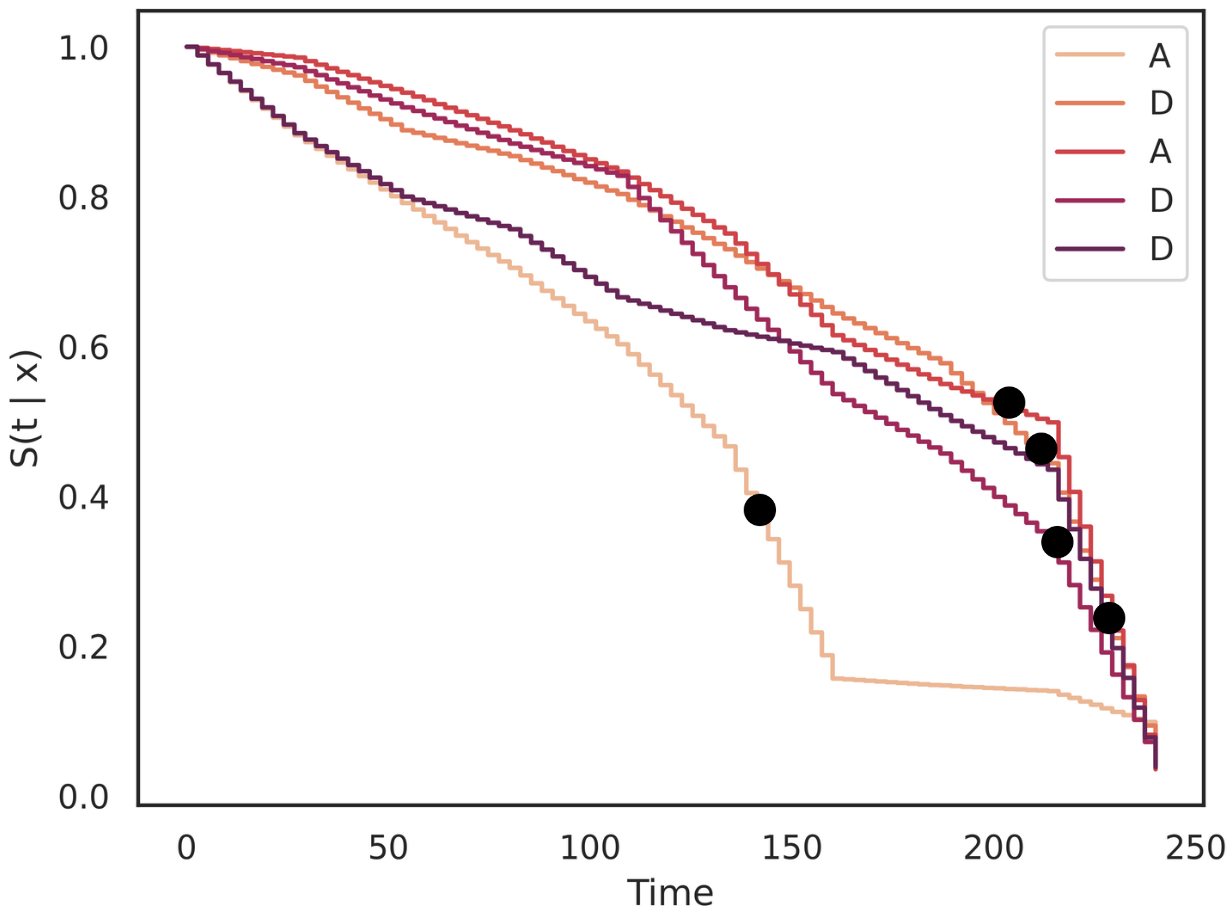}
            \caption[]%
            {{\small Survival curve (estimate of survival probability over time) for five random samples from MIMIC-IV using both static and time-series features with DySurv}}
            \label{fig:mean and std of net44}
        \end{subfigure}
        \caption
        {\small The set of survival curves for the MIMIC-IV ICU EHR dataset as generated by DySurv shows extrapolation of risk across different patients as compared to static and time-series feature sets. Dots correspond to true event times which were predicted correctly by DySurv.} 
        \label{fig:survival_curves_MIMIC}
    \end{figure}

\noindent In survival analysis, it is also vital to show the performance of the models in creating survival curves, or estimates of the survival probabilities over time for different patients. Whereas traditional survival analysis models from statistics rely on risk sets and computing survival estimates for a population, an advantage of deep learning models is that the task of survival estimation can apply to an individual patient sample. We provide survival curves for a random group of five samples/patients (with different event times) across all datasets that show a clear separation of risk as can be seen in Figure \ref{fig:survival_curves}. For MIMIC-IV, the data input consists of both static and time-series data and we provide survival curve results for both scenarios in Figure \ref{fig:survival_curves_MIMIC}. \\

\begin{figure}[H]
        \centering
        \begin{subfigure}[b]{0.46\textwidth}   
            \centering 
            \includegraphics[width=\textwidth]{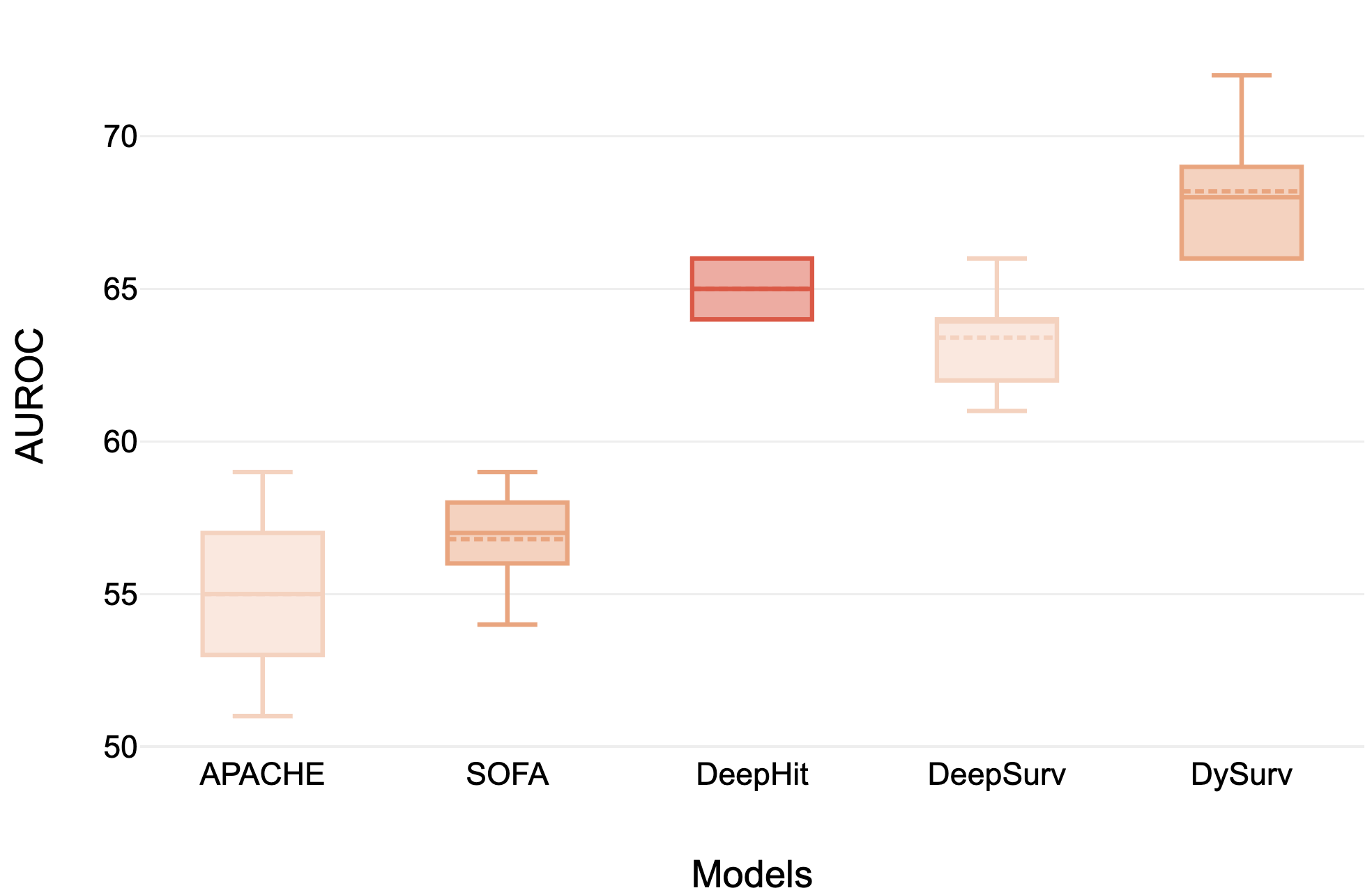}
            \caption[]%
            {{\small}}    
            \label{fig:mean and std of net34}
        \end{subfigure}
        \hfill
        \begin{subfigure}[b]{0.46\textwidth}   
            \centering 
            \includegraphics[width=\textwidth]{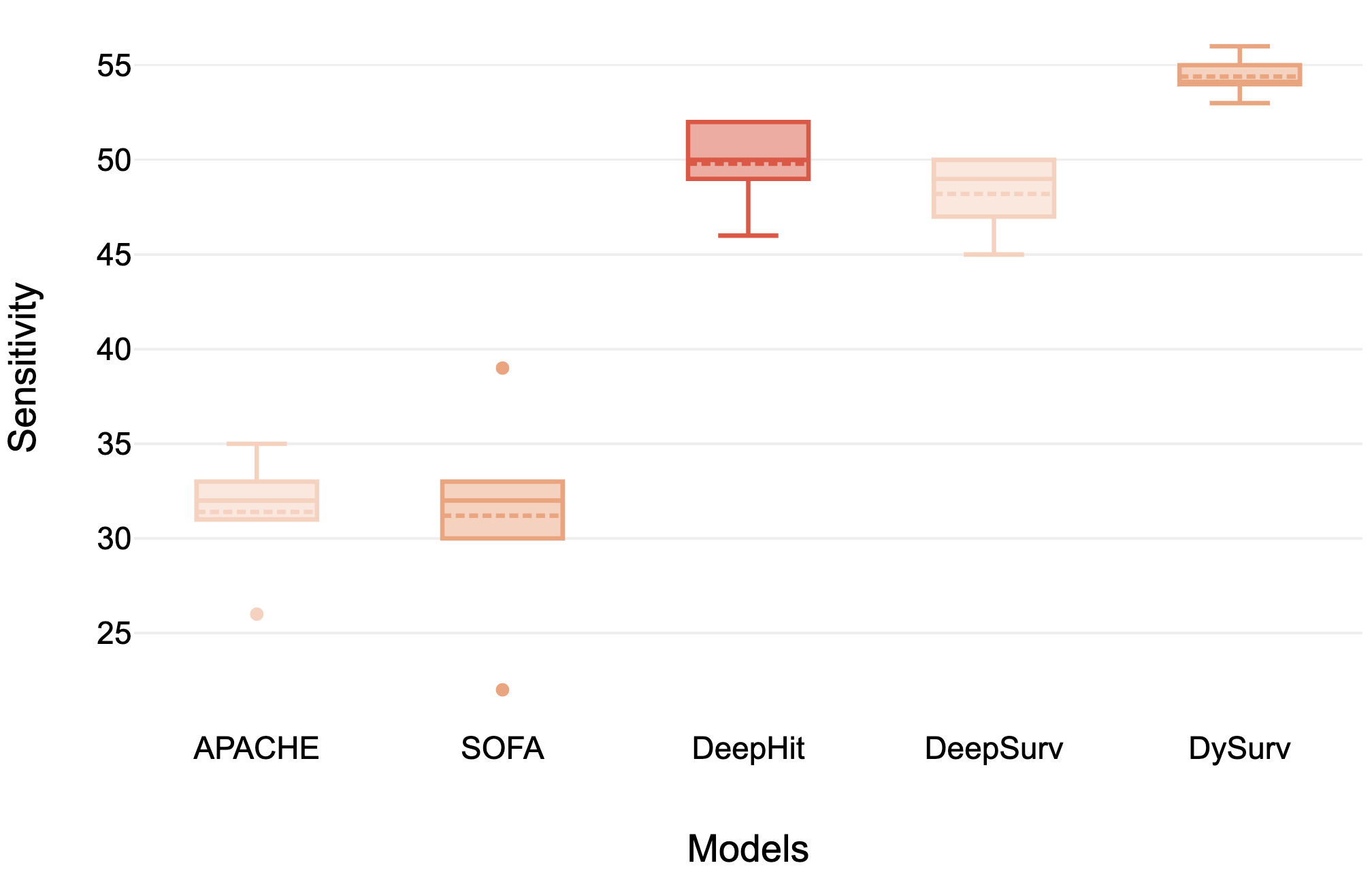}
           \caption[]%
            {{\small}}
            \label{fig:mean and std of net44}
        \end{subfigure}
        \vfill
        \begin{subfigure}[b]{0.46\textwidth}   
            \centering 
            \includegraphics[width=\textwidth]{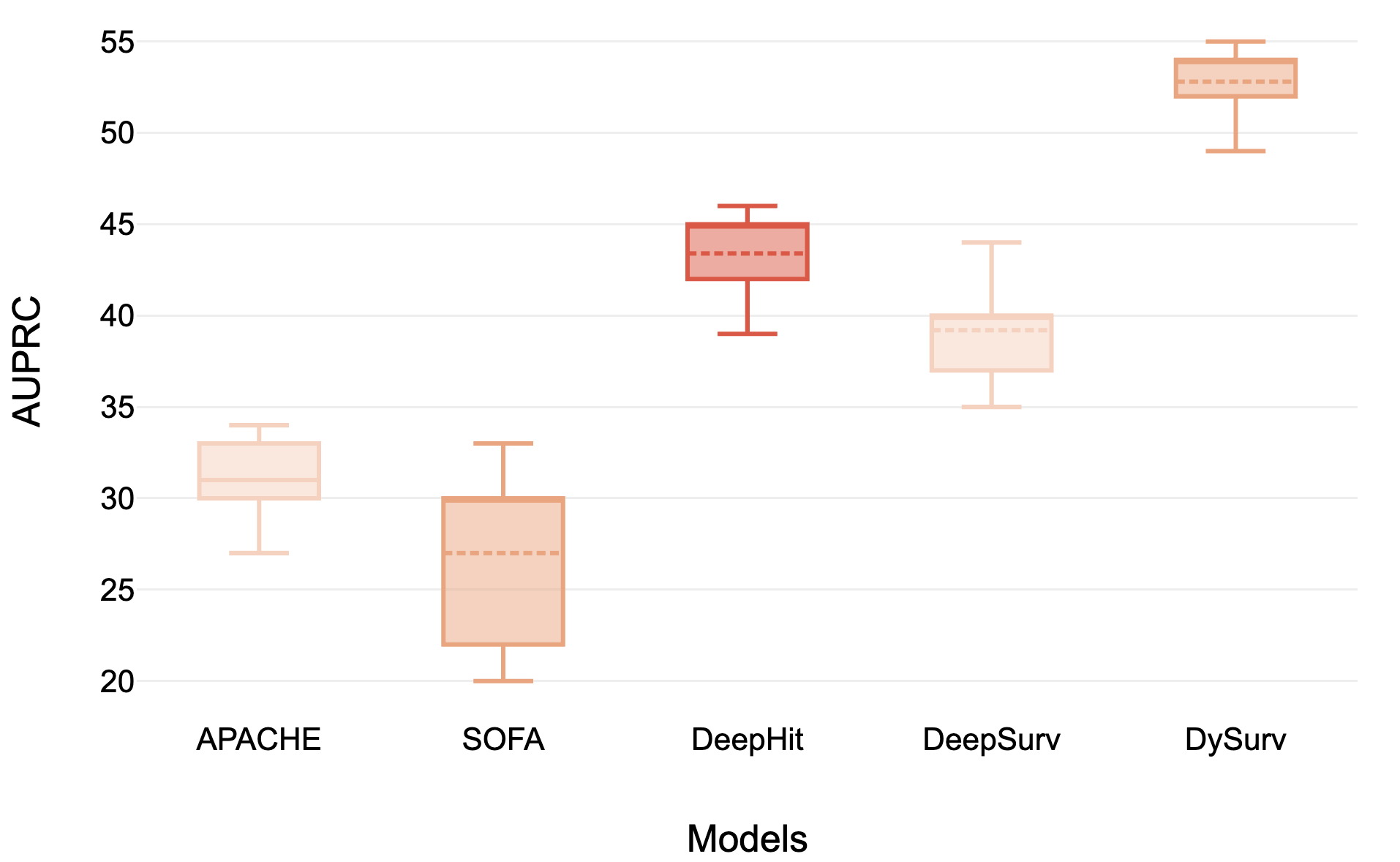}
           \caption[]%
            {{\small}}
            \label{fig:mean and std of net44}
        \end{subfigure}
        \caption
        {\small Comparison of DySurv at 24-hour prediction with existing ICU survival scores and deep learning survival models on MIMIC-IV using (a) AUROC and (b) Sensitivity and c) AUPRC. APACHE IV was only able to be retrieved from MIMIC-IV and SOFA score was calculated from the eICU dataset.} 
        \label{fig:APACHE_SOFA}
    \end{figure}

\noindent Upon deployment of a trained model, DySurv can generate risk estimates through time for each patient while using their history of observations. We do not rely on landmarking methods or a specific pre-defined time for risk prediction as the scores are simultaneously issued across the entire time interval. For data pre-processing purposes, a time horizon is selected corresponding to 24 hours with a 72-hour timespan for the LSTM in the case of MIMIC-IV and eICU since ICU risk assessments often use information over 72 hours for the next 24-hour risk prediction \cite{yu2014comparison}. We also compare DySurv for 24-hour risk prediction to existing ICU survival scores in practice like APACHE IV and SOFA scores which can be seen in Figure \ref{fig:APACHE_SOFA}.\\

\section{CONCLUSIONS}
\noindent The first set of results from Table \ref{tab:Result} relate to applying DySurv only on static data from several benchmark datasets of varying sizes. We see that for the vast majority of these benchmarks, DySurv outperforms both standard statistical as well as deep learning alternatives across all metrics except for METABRIC and NWTCO on the concordance where DeepHit tends to perform slightly better. This is probably due to the implementation of the biased ranking loss mentioned earlier that aids in having better discriminative performance as measured by the concordance metric but that is not reflected as measured by the other two metrics. Similar behaviour for DeepHit has been observed in another study by \cite{kvamme2019time}. We also see that the non-VAE implementation of the logistic hazard performs much worse than DySurv across all experiments, thereby strengthening the idea that adding variational inference to the logistic hazard can aid in learning the survival task. This improvement occurs despite having the additional task of reconstruction now. The predictive advantage comes from the identification of lower dimensional latent vectors used in the survival task instead of the raw features directly. Furthermore, on very large synthetic datasets, such as sac3 and sac\_admin5, DySurv performs better due to having a greater amount of data to learn from. A limitation of the other benchmark datasets is their relatively small size may constrain DySurv from learning its optimal parameters to provide better survival prediction. \\

\noindent While there are previous models that have attempted to use autoencoders for survival analysis, such as \cite{kim2020improved} and \cite{tong2020deep}, they have not explored variational inference extensively. These models also rely on optimisation of the Cox partial log-likelihood loss, hence being restricted by the proportionality assumption and they do not account for dynamic time-series or time-varying features in the input. Furthermore, work by \cite{kim2020improved} suggests that the VAE model's learned compact latent representation directly aids in the improved performance of the Cox model. This intuition is precisely what we have also seen in our results, albeit in a larger, more flexible, and more complex scenario of time-series ICU risk prediction with direct joint distribution estimation. The concatenation autoencoder from \cite{tong2020deep} is not even compatible with static data, by far the most common modality in survival analysis, thereby limiting its relevance significantly. DySurv addresses all these limitations and provides a flexible solution to dynamic survival analysis with deep learning. \\

\noindent As for MIMIC-IV and eICU results, we see that even with only static data in the input, DySurv manages to outperform other survival analysis models. When time-series data are added to the input in a multi-modal fashion, it can help the model improve its performance and outperform both CoxTime and Dynamic-DeepHit across all metrics. Figure 3 shows examples of projected survival trajectories using DySurv in both the static-only and static-and-time-series feature settings. We also see that, as expected in the hospitalised setting, the survival of the patients significantly changes in the last few days in the ICU, starting to drop a few days before death. Previous work has shown that earlier discharge times in the ICU correspond to higher survival rates. This suggests that identifying the period when survival rates drop dramatically in the ICU can help target earlier treatment for those most at risk \cite{simchen2004survival}. We assume that patients will stay in the ICU for a maximum of 10 days and while that is generally true, there could be outliers and differences in ICU stay duration distributions. We suspect this would not severely hinder the generalizability of DySurv to those scenarios as that would just mean an expansion of the time component for the survival trajectory predictions. \\

\noindent From Figure 4 we can see how DySurv outperforms existing in-use systems for patient risk prognosis in the ICU across a variety of metrics. We acknowledge the limitation of the results for generalisability from using different data subsets. The model is highly dependent on the quality and completeness of time-series data. While forward and backward filling has been commonly employed and verified with clinical collaborators \cite{nordmark2021practice, rocheteau2021temporal}, this approach could introduce biases, particularly if the missingness is non-random or systematically tied to patient outcomes. Furthermore, we used pre-defined time intervals (72-hour windows) which may not fully capture the dynamic variability in patient trajectories, particularly for patients with more erratic health progressions. This time interval, however, can be manually adjusted depending on the identified application. We also acknowledge potential overfitting to the benchmark datasets, especially given the relatively small size of some of these datasets. This could hinder generalization to more diverse populations or clinical settings where patient characteristics and ICU protocols differ significantly. We have found that careful hyperparameter optimisation can aid in addressing this challenge when learning on different datasets. \\

\noindent The generalizability of DySurv is also constrained by the assumptions embedded in the model architecture, despite being nonparametric. While DySurv does not assume proportional hazards, it still relies on the assumptions inherent in its logistic hazards model producing discrete estimates of the survival probabilities. Additionally, as deep learning models are inherently opaque, their extensions to healthcare settings come with concerns of trustworthiness and interpretability. To address this limitation and further verify that DySurv is learning relevant clinical patterns from the patient data, we used permutation importance to obtain the top 10 predictive features for DySurv on MIMIC-IV and eICU. These results are in line with relevant clinical literature and practice and a discussion can be found in Section G in the Supplementary. \\

\noindent In this paper, we present a novel dynamic risk prediction model for survival analysis based on deep learning in survival analysis paradigms. We work collaboratively with intensive care physicians and demonstrate interdisciplinary pathways to the implementation of deep learning for survival analysis methods in the ICU as shown in Figure 2 in the Supplementary. Our method builds on a combination of previous work including Dynamic-DeepHit by leveraging direct learning of the joint distribution of the first event time and the event through log-likelihood optimisation with logistic hazards. Theoretically, this approach is an alternative to the risk log-likelihood loss function of Dynamic-DeepHit itself, which does not use a ranking loss for biased inflation of concordance results. Our DySurv model is capable of learning from complex EHR ICU time-series data and extracting lower-dimensional latent representations that can be useful for learning the survival task while also balancing reconstruction. As the model has been difficult to train due to loss instabilities and sensitivity to hyperparameter selection, future work can explore including a regularization component to the loss terms. By using the underlying latent distribution, we can directly model an alternative to the survival distribution like the Weibull distribution instead of using a Gaussian intermediate. \\


\section{Funding Statement}
\noindent This work was supported by the Rhodes Trust and the EPSRC Centre for Doctoral Training in Health Data Science grant (EP/S02428X/1). \\

\section{Competing Interests Statement}
\noindent The authors have no competing interests to declare. \\

\section{Contributorship Statement}
\noindent MM, PW, and TZ have all contributed to the conception or design of the work; or the acquisition, analysis, or interpretation of data for the work. MM has extracted and pre-processed the data with guidance from PW and TZ. MM has developed the method with guidance from TZ, designed the experiments, validated the results, and summarised the findings. MM, PW, and TZ drafted the work or reviewed it critically for important intellectual content. MM provided the initial draft and implemented the subsequent changes in writing. MM, PW, and TZ approved the final version for publication. All authors agree to be accountable for all aspects of the work in ensuring that questions related to the accuracy or integrity of any part of the work are appropriately investigated and resolved. 

\section{DATA AVAILABILITY}
\noindent The data is available from public access requests for eICU and MIMIC-IV. Accessing the data requires ethics module training and certification. It can be obtained at the following links: \url{https://physionet.org/content/eicu-crd/2.0/} and \url{https://physionet.org/content/mimiciv/2.0/}. \\



\ifCLASSOPTIONcaptionsoff
  \newpage
\fi

\bibliographystyle{vancouver} 
\bibliography{mortality_only}

\end{document}